\documentclass{ws-ijait}
\usepackage[utf8]{inputenc}
\usepackage{graphicx}

\usepackage{url}

\begin{document}

\markboth{Ale\v{s} Hor\'ak, V\'it Baisa, Adam Rambousek, V\'it Suchomel}{Semi-automatic Tools for Multilingual Terminology Thesaurus}
\title{A New Approach for Semi-automatic Building and Extending a~Multilingual Terminology Thesaurus\footnote{Preprint of an article submitted for consideration in International Journal on Artificial Intelligence Tools © 2019 copyright World Scientific Publishing Company\newline https://www.worldscientific.com/worldscinet/ijait }}

\author{Ale\v{s} Hor\'ak}

\address{Natural Language Processing Centre\\
Faculty of Informatics, Masaryk University\\
Botanicka 68a, 602 00 Brno, Czech Republic\\
\texttt{hales@fi.muni.cz}
}

\author{V\'it Baisa}

\address{Natural Language Processing Centre\\
Faculty of Informatics, Masaryk University\\
Botanicka 68a, 602 00 Brno, Czech Republic\\
\texttt{xbaisa@fi.muni.cz}
}

\author{Adam Rambousek}
\address{Natural Language Processing Centre\\
Faculty of Informatics, Masaryk University\\
Botanicka 68a, 602 00 Brno, Czech Republic\\
\texttt{rambousek@fi.muni.cz}
}

\author{V\'it Suchomel}
\address{Natural Language Processing Centre\\
Faculty of Informatics, Masaryk University\\
Botanicka 68a, 602 00 Brno, Czech Republic\\
\texttt{xsuchom2@fi.muni.cz}
}
\maketitle


\begin{abstract}
This paper describes a new system for semi-automatically building, extending and managing a terminological thesaurus---a multilingual terminology dictionary enriched with relationships between the terms themselves to form a thesaurus. The system allows to radically enhance the workflow of current terminology expert groups, where most of the editing decisions still come from introspection. The presented system supplements the lexicographic process with natural language processing techniques, which are seamlessly integrated to the thesaurus editing environment. The system's methodology and the resulting thesaurus are closely connected to new domain corpora in the six languages involved. They are used for term usage examples as well as for the automatic extraction of new candidate terms. The terminological thesaurus is now accessible via a web-based application, which
a) presents rich detailed information on each term,
b) visualizes term relations, and
c) displays real-life usage examples of the term in the domain-related documents and in the context-based similar terms. Furthermore, the specialized corpora are used to detect candidate translations of terms from the central language (Czech) to the other languages (English, French, German, Russian and Slovak) as well as to detect broader Czech terms, which help to place new terms in the actual thesaurus hierarchy.

This project has been realized as a terminological thesaurus of land surveying,
but the presented tools and methodology are reusable for other terminology domains.
\end{abstract}


\section{Introduction}
\label{sec:introduction}

Specialists in any branch inevitably rely on domain-specific vocabulary as a basis for sharing exact terminology among professionals. Such detailed domain terminology cannot be included in general language dictionaries, which is why specialized terminology dictionaries are being built and managed. With the need to share information unambiguously in different languages, terminology dictionaries often link original terms to their translations. The taxonomic ordering of the terminology is described by means of term relations such as synonymy or hypernymy/hyponymy.\footnote{\textit{Synonymy} is usually used in a ``weakened'' form as \textit{near synonymy} or the \textit{see also} relation. \textit{Hypernymy/hyponymy} refers to \textit{broader/narrower terms}.} In the system presented in this paper, information about the terms is described and visualized in a way that helps the readers (both specialists and the general public) to understand the meaning of the term and its usage in contexts.

Any human language is continuously evolving---new words and terms appear while usage and meanings evolve. This evolution is even more noticeable in specialized vocabularies.\cite{fischer1998lexical} A system for terminology thesauri thus needs to cope with frequent updates of the data. In this article, we present the details of a thesaurus development system which prepares the underlying information for updates automatically, and displays them during the entry editing process for terminologists' authorization and completion.

The Natural Language Processing Centre (NLP Centre) at the Faculty of Informatics, Masaryk University, in cooperation with the Czech Office for Surveying, Mapping and Cadastre (CUZK) has developed a new system for building and extending a specialized terminological thesaurus for the domain of land surveying and land cadastre, which we refer to as TeZK.\footnote{In Czech "Tezaurus pro obor zem\v{e}m\v{e}\v{r}ictv\'{i} a katastru nemovitost\'{i}" (Thesaurus for the field of land surveying and land cadastre).} The TeZK project consists of several tightly interconnected parts---a web-based application to create, edit, browse and visualize the terminological thesaurus, and a set of tools to build large corpora of domain oriented documents which allows for the detection of newly emerging terms, or terms missing from the thesaurus. General tools already developed by the NLP Centre for corpus building and term extraction and a platform for dictionary applications are utilized here alongside newly developed tools for extracting candidate translations and for identifying candidate broader terms.\footnote{In lexicography a ``broader term'' denotes a ``hypernym,'' we use these two denotations interchangeably in the text.} These have been developed on top of the general corpus tools. During the TeZK project, we enhanced the existing corpus tools so as to support comparable multilingual corpora. We also developed a new thesaurus web application (not limited to single domains) with new methods that interconnect the domain corpus with the terminological thesaurus. Each term in the thesaurus is supplemented with knowledge-rich context information---a term explanation, its relationships to other terms, usage examples, term translations, or specific links to related resources, such as e.g. the corresponding legislative.\cite{meyer2001extracting} In this sense, the system offers a radical improvement of the work of terminology expert groups who are in charge of organizing, constructing and managing the official terminology in their respective area.

After evaluating the TeZK project, the Czech e-Government Office decided to use it as a basis for a new official Czech registry of terminological thesauri, currently in the early development phase. In the follow-up project, the terminological thesaurus system is being updated to support easy and user friendly deployment at any organization, with the possibility to customize work processes based on specific organization requirements. Furthermore, each instance of the terminological thesaurus system will share selected data with the central registry and all other terminological thesauri.

This article is structured as follows. Section~\ref{sec:related} contains an overview of related work regarding systems for the advanced building of terminology thesauri. In Section~\ref{sec:specialized}, we present the process of creating a large specialized domain corpus with the functionality of extracting new candidate terms in the selected domain. In Section~\ref{sec:application}, we describe the multiplatform web-based editor and browser application which is based on DEB, an existing dictionary writing platform now enhanced with new functions for the thesauri building system. Although the TeZK project aimed at building and managing a terminological thesaurus of the land surveying domain, our newly developed tools may be re-used for any other domain dictionary, which would facilitate the sharing of information and stimulate a general awareness of a selected domain. We also introduce the linking capabilities of the terminological thesaurus data in the system as related to Linked Open Data methodology. The final section provides an overview of its usage and outlines some possible areas for future work.

\section{Related work}
\label{sec:related}

The idea of developing a specialized system for creating and sharing terminological thesaurus content is not new, as several tools for dictionary management are available, both free and commercial, such as Lexique Pro\footnote{\url{http://www.lexiquepro.com/}} or tlTerm\footnote{\url{http://tshwanedje.com/terminology/}}. However, these applications concentrate on features for dictionary compilation and presentation. We, on the other hand, are aiming at enhancing the whole methodological process by adding knowledge-rich context information for each term and automatic candidate information about new terms, their taxonomical relations and translations. For the same reason, we do not reuse any complete existing tool for building the TeZK terminological thesaurus, only include software tools developed for single tasks and enhance them to form the complete system.

Our thesaurus management system exploits our previous experience in designing and developing several applications for terminological thesaurus creation, such as the Multilingual Glossary of Fine Art Terminology with the Faculty of Fine Arts, Brno University of Technology, or the Czech-English Dictionary of Ethnological Terminology with the National Institute of Folk Culture.\footnote{\url{http://www.nulk.cz/}} The current terminology thesaurus project considerably extends the features developed in the other projects.

\subsection{Data visualization}
Previous research showed that visualization and rich information significantly help dictionary users in understanding the terms and context. Graphic representations of the relationships between terms proved indispensable to users, as is the case in the DiCoInfo Visuel project\cite{ROBICHAUD12.1096} and in EcoLexicon\cite{Faber2014,leon2016pattern,leon2018ecolexicon}. We thus decided to add options that would visualize the term relationships, and a tree representation of hyponymy/hypernymy relations. Several studies also found that rich information is very helpful for students and translators in the field. Marshman's case study concluded that ``Studying terms in context can also help clarify differences between concepts expressed by polysemous terms.''\cite{marshman2014enriching}

\subsection{Term extraction}
Automatic term extraction provides candidate term lists as support materials for subject field experts. Corpora-based term extraction uses lexical collocability measures combined with other evidence and approaches. For example, the TExSIS system\cite{macken2013texsis} proposes an algorithm that works with bilingual parallel corpora: it finds aligned chunks of texts (using statistical word alignment) and extracts the translated terms using statistical filters. Even though the results look promising, they are limited to language pairs for which precisely aligned corpora are available.

Another approach was proposed by García-Silva et al.\cite{garciasilva2015} to automatically construct domain ontologies based on social tagging systems. Such systems (also called folksonomies) allow users to publicly share webpage links and add labels or tags used to categorize the content. However, this approach is not useful for the TeZK project because of the lack of data for the specialized domain.

We decided to integrate a technique (described in Section~\ref{sec:specialized}) that has been successfully included in the Sketch Engine corpus manager and verified in a terminology extraction project from patent data of the World Intellectual Property Organisation.\cite{1181590}

\subsection{Term translation}
The TeZK system offers a list of candidate term translations based on specifically prepared domain corpora\footnote{Text corpora with documents devoted to a selected field or problem domain, see Section~\ref{sec:specialized} for further details.}. The idea of extracting translation candidates from comparable corpora\footnote{Comparable corpora (as opposed to ``parallel corpora'') are text corpora in different languages, whose documents talk about the same topics but are not direct translations of each other.} has been studied by Morin et al.\cite{Morin:2008:BBU:1839478.1839479}, who have shown that the quality of comparable corpora might alleviate the data sparsity problem. This is also the case of the TeZK system where the selected domain is rather limited. Sadat et al.\cite{Sadat:2003:BTA:1075178.1075201} proposed a system (for the Japanese-English language pair) which first extracts possible translation pairs and then filters out non-promising candidates using linguistic rules. Gu et al.\cite{Gu2014} used a similar approach to discover semantically similar sentences, however only within a single language.

Daille and Morin\cite{Daille:2005:FTE:2145899.2145979} used contexts for aligning possible translation candidates of previously extracted monolingual multiword expressions from French-English technical documents. 
Lee et al.\cite{Lee:2010:EHM:1944566.1944639} used an EM-based\footnote{Expectation Maximization} algorithm for the extraction which required an alignment of comparable documents prior to the actual candidate extraction. They demonstrated a language independent approach on English-Chinese and English-Malay language pairs. In one part of the extraction procedure, they used co-occurrence statistics. Sorg and Cimiano\cite{Sorg201226} proposed multi-lingual concept linking with the help of explicit semantic analysis, using Wikipedia categorization and cross-language links.

\subsection{Semantic relations}
Another feature of the TeZK system is the extraction of semantic lexical relations, particularly hypernyms and hyponyms, i.e. broader and narrower terms. This technique is generally used for augmenting or verifying existing lexicons and for identifying semantically related terms as proposed by Hearst.\cite{Hearst:1992:AAH:992133.992154} 

In the TeZK project, hypernym candidate identification is used when adding a new term to the ontology or taxonomy built within the system. Hearst identifies the lexico-syntactic patterns by bootstrapping from manually discovered patterns or existing lexicons, and deriving new rules from common syntactic environments. Hearst also argues that this technique does not work well for English meronymy/holonymy. 

Snow et al.\cite{snow2004learning} propose learning the patterns automatically via a logistic regression classifier trained over texts containing hypernym/hyponym word pairs from the WordNet semantic network. Banko et al.\cite{banko2008tradeoffs} presented an open information extraction method, which is based on extracting occurrences of different relations using a small set of general relation patterns common to all kinds of relations and then deciding the relations by a CRF-based\footnote{Conditional Random Fields} unsupervised extraction. The best recall and precision was achieved by a combination of supervised and unsupervised approaches. Arnold and Rahm\cite{arnold2014} proposed an algorithm to extract semantic relations from Wikipedia corpora which may also prove useful for lexicon enhancement; however, the algorithm was tested only for English data. 

A case study by Lefever et al.\cite{lefever2014hypoterm} describes the HypoTerm system for hypernym detection in Dutch and English. The paper evaluated multiple approaches for relationship detection (pattern-based, morpho-syntactic analysis, statistical, WordNet-based) and discovered that the pattern-based approach provides the highest precision scores.

The TeZK system follows the pattern-based approach for the sake of higher precision and ease of maintenance within the thesaurus system. With the growth of semantic web technologies, machine learning from semantic web data may prove better results, as discussed by Rettinger et al.\cite{Rettinger2012}

\section{Specialized Corpora and Term Extraction}
\label{sec:specialized}

The advanced automatic functionality of the TeZK system relies on a set of large text corpora in six predetermined languages containing domain-specific texts, i.e. mostly technical or popular text devoted (in this case) to topics related to land surveying, geodesy and land cadastre in general. Our previously developed web crawler, SpiderLing\cite{suchomel2012efficient},  and tools for web text cleaning\cite{justext} were used to build new domain corpora from publicly available online resources in Czech, English, German, French, Russian and Slovak. 
The process of constructing the specialized domain corpora started with the ``pivot'' language, Czech. Firstly, a set of main websites related to the land surveying, the cadastre of real estates, and related topics was listed. See Table \ref{tab:sources} for details of the primary sources for the Czech corpus. We use standard corpus size measures, such as the number of documents and the number of tokens, i.e. text units like words, punctuation characters and structure tags.

\begin{table}[ht]
\tbl{Primary website sources for the Czech domain corpus. Tokens are particular word, punctuation or structure tag occurrences. Unique documents refer to the number of documents after removing (near) duplicate documents. Unique tokens describe the vocabulary richness of the respective source.}
{\begin{tabular}{lrrrr}
\hline
\textbf{Website} & \textbf{Docs} & \textbf{Tokens} & \textbf{U docs} & \textbf{U tokens} \\ \hline
www.cuzk.cz & 16,405 & 3,137,795 & 15,289 & 340,943 \\
www.vugtk.cz & 4,659 & 6,419,950 & 3,212 & 4,386,238 \\
csgk.fce.vutbr.cz & 241 & 77,255 & 198 & 58,561 \\
www.kgk.cz & 417 & 44,814 & 414 & 29,890 \\
www.sfdp.cz & 192 & 35,287 & 106 & 11,279 \\
www.czechmaps.cz & 94 & 108,506 & 90 & 98,914 \\
www.zememeric.cz & 8,634 & 6,100,751 & 6,200 & 2,638,308  \\ \hline
\end{tabular}}
\label{tab:sources}
\end{table}

\noindent
Secondly, a broader set of documents from a large set of websites  was obtained by our WebBootCat tool\cite{webbootcat} based on the ``term content'' (see below) of the primary websites. Table~\ref{tab:crawled_languages} presents the number of acquired documents and tokens for all languages covered in the project. This method needs a set of seed words to search the web for relevant documents. For the seed word sets, we have used the main domain terms obtained from the publicly available terminology dictionary.\cite{vugtk} The representativeness of the created corpora and their thematic coverage of the selected areas can be seen in Table~\ref{tab:subdomains} with details about the document and token distribution among different sub-topics (as divided in the authoritative terminology dictionary). Non-textual and low quality content was automatically identified and removed from the downloaded documents utilizing our JusText tool.\cite{justext} Finally, duplicate documents or paragraphs were purged with our Onion tool.

\begin{table}[ht]
\tbl{Statistics detailing the web-crawled domain corpora for the six languages.}
{\begin{tabular}{lrrr}
\hline
\textbf{Language} & \textbf{Documents} & \textbf{Tokens} & \textbf{Web domains} \\ \hline
English & 8,149 & 40,225,064 & 4,946 \\
French & 5,326 & 15,789,761 & 3,291 \\
German & 3,373 & 9,744,313 & 2,220 \\
Russian & 2,914 & 19,015,734 & 1,770 \\
Slovak & 2,943 & 10,252,449 & 1,528 \\
Czech & 27,389 & 12,689,548 & 1,061 \\ \hline
\end{tabular}}
\label{tab:crawled_languages}
\end{table}

\begin{table*}
\tbl{Distribution of sub-topics of the resulting corpora (in percentages), per language, by the number of documents and by the number of tokens. Cadastre, cartography and geodesy are the most frequently represented topics.}
{\begin{tabular}{lrrrrrr}
\hline
language & \multicolumn{2}{c}{English} & \multicolumn{2}{c}{French} & \multicolumn{2}{c}{German} \\ \hline
sub-topic & docs & tokens & docs & tokens & docs & tokens \\  \hline
cadastre & 9.7 & 10.4 & 6.2 & 6.8 & 13.9 & 10.1 \\
cartography & 17.6 & 20.2 & 16.5 & 15.9 & 26.0 & 22.3 \\
engineering surveying & 3.6 & 11.2 & 7.8 & 6.5 & 12.1 & 9.5 \\
theory of errors & 5.3 & 2.6 & 4.4 & 2.9 & 3.6 & 9.0 \\
geodesy & 20.1 & 19.4 & 25.4 & 24.2 & 1.6 & 3.7  \\
geoinformation & 13.0 & 6.1 & 5.7 & 12.7 & 7.8 & 7.9  \\
GPS system & 7.7 & 4.1 & 5.0 & 5.2 & 0.8 & 0.6  \\
instrumental technology & 7.2 & 5.1 & 4.3 & 2.2 & 2.2 & 9.1  \\
mapping & 6.8 & 10.0 & 9.8 & 11.5 & 21.4 & 14.6 \\
metrology & 6.2 & 8.4 & 4.2 & 5.5 & 6.5 & 7.1  \\
photogrammetry & 3.0 & 2.4 & 10.8 & 6.5 & 4.1 & 6.0  \\ \hline
 & \multicolumn{2}{c}{Russian} & \multicolumn{2}{c}{Slovak} & \multicolumn{2}{c}{Czech} \\ \hline
cadastre & 6.2 & 3.4 & 15.2 & 10.7 & 15.4 & 14.5 \\
cartography &  11.4 & 9.1 & 16.0 & 21.0 & 21.7 & 21.0 \\
engineering surveying &  6.3 & 7.8 & 7.1 & 5.0 & 6.7 & 4.5 \\
theory of errors &  7.4 & 10.5 & 1.5 & 2.0 & 4.5 & 4.1 \\
geodesy &  18.2 & 27.1 & 21.0 & 15.4 & 11.2 & 9.0 \\
geoinformation &  16.9 & 14.2 & 6.0 & 5.9 & 3.3 & 10.3 \\
GPS system & 2.8 & 1.0 & 3.3 & 2.5 & 6.9 & 4.1 \\
instrumental technology & 4.8 & 6.1 & 5.5 & 1.9 & 6.7 & 3.2 \\
mapping & 7.3 & 4.9 & 12.1 & 18.6 & 12.6 & 13.5 \\
metrology & 12.6 & 9.0 & 2.5 & 2.8 & 8.5 & 11.5 \\
photogrammetry & 6.2 & 6.7 & 9.8 & 14.2 & 2.5 & 4.3 \\ \hline
\end{tabular}}
\label{tab:subdomains}
\end{table*}

\subsection{Automatic Extraction of New Candidate Terms}
\label{subsec:termextraction}

These domain corpora offer a sufficient basis for the intelligent functionality of the TeZK system. All the corpora can be continuously updated (extended) by adding new documents (or websites), which will undergo the same processing pipeline as described above, i.e. text extraction, deduplication and tokenization. The first function allows us to identify ``candidate terms'', i.e. proposals to be checked by experts in the subject field and easily added to the thesaurus. The candidate terms were extracted from the domain corpora using a hybrid approach: a) the first step is a linguistically motivated rule-based extraction of noun phrases and other term patterns~\cite{jakubivcek2009mining}, and b) the second step lies in sorting all these noun phrases by relevance which is computed by comparing the relative frequency of each noun phrase in the given domain corpus with its relative frequency in a (general) reference corpus~(\cite{compare,simplemaths}).

The first step, i.e. identification of phrases in the corpus text, which could form a (complex) term, is based on a predefined (language-dependent) set of patterns denoted as a ``term grammar.'' These patterns are expressed in the form of Corpus Query Language (CQL) and describe phrases such as a (complex) noun phrase (e.g. ``\emph{digital photogrammetric workstation}'') or a combination of a noun phrase and a prepositional phrase (e.g. ``\emph{parallactic figure with an auxiliary base}'').

\noindent
In the second step, the resulting ``term rank'' of each identified phrase is determined by the formula 
$$\mathit{rank}(\mathit{term\_candidate}) = \frac{f + n}{f_{\mathit{ref}} + n},$$
where $f$ is the domain corpus relative frequency of a given candidate term, and $f_{\mathit{ref}}$ its relative frequency in a reference corpus. The parameter $n$ (called simple math) can be used to fine-tune the results based on the size of the analyzed corpora and on user's preferences. High values of $n$ cause the algorithm to prefer more frequent phrases and vice versa. In the default setup the value of $n=1$ is chosen. This approach allows to adapt the term extraction technique to the specific language data in cases where standard statistical methods (e.g. mutual  information (MI) score,  Log-Likelihood, or Fisher’s Exact Test) fail due to their assumption that the language phenomena are independent of each other -- see~\cite{simplemaths} for a detailed explanation.

For each language, the respective corpus of the TenTen corpus family was used as a reference corpus.\cite{tenten} The TenTen corpus family contains very large general language corpora\footnote{The sizes of the TenTen corpora range from billions of words to tens of billions of words, ergo $10^{10}$ words. The TenTen corpus family currently covers 31 languages.} built from web.

\subsection{Automatic Term Relations Identification}
\label{subsec:hypernym}

The methodology of the TeZK systems aims at continuous amendments of the thesaurus taxonomy using new terms. The term inclusion process is supported by two other (semi)automatic techniques: the identification of candidate hypernyms/broader terms and the candidate term translations. The technique of broader terms identification relies on two methods of automatic hypernym extraction: a pattern extraction from a domain corpus and a term similarity based approach.

Within the \textit{pattern extraction method}, the specialized domain corpus (of the pivot language) is filtered\footnote{The queries are evaluated via the concordance API of Sketch Engine\cite{kilgarriff2014sketch} with the patterns specified in the formal Corpus Query Language (CQL).} to obtain a list of possible hypernym candidates, which are then ordered using the logDice similarity score:

$$\mathit{logDice}(t_1, t_2) = \log_2 \bigg(\frac{2 f_{t_1, t_2}}{f_{t_1} + f_{t_2}}\bigg),$$

\noindent
where $f_{t_1,t2}$ is the number of co-occurrences of terms $t_1$ and $t_2$.

The number of possible patterns can be generally extended without limitations. The TeZK system uses three of the most productive patterns:
\begin{itemize}
\item Pattern 1: \emph{The hyponym} + \emph{is/are} + \emph{the hypernym},
\item Pattern 2: \emph{The hyponym} + \emph{and/or another\discretionary{/}{/}{/}other/similar} + \emph{the hypernym}, 
\item Pattern 3: \emph{The hyponym} + \emph{is/are a kind\discretionary{/}{/}{/}type\discretionary{/}{/}{/}part\discretionary{/}{/}{/}example\discretionary{/}{/}{/}way of} + \emph{the hypernym}.
\end{itemize}

\noindent
Although the accuracy of Pattern 1 and Pattern 2 queries is above 50\%
, not all successfully extracted hypernym pairs are suitable for the particular term database. For instance, some identified hypernym terms are too general to be included in the thesaurus or, vice versa, explicit hyponyms are particular instances, which are not to be included in the definitions according to the editor's decision.
Another approach to finding hypernyms of a term involves searching the current term database and identifying \textit{lexically similar terms}, e.g. ``Cartesian coordinate system'' and ``coordinate system''. The most similar terms are expected to be good generalizations of the term, and thus either good hypernym candidates or synonym terms, which help to identify a common hypernym. The lexical similarity measure between two terms is based on the Jaccard distance of bigrams of characters with a threshold of 0.5:

$$lexsim(t_1, t_2) = \frac{|t_1 \mathit{bigrams} \cap t_2 \mathit{bigrams}|}{|t_1 \mathit{bigrams} \cup t_2 \mathit{bigrams}|}$$

\noindent
Both these methods are combined in the system and the best candidates for hypernyms are available for the terminologists in the user interface in the form of a shortcut select box (see Section~\ref{subsec:editing}). The final decision as to which candidate to select or whether to input the hypernym manually is still left to a terminologist.
During the development of the relation extraction module, we have evaluated 7~different patterns. We have evaluated 50 candidate relation pairs per each of the extraction patterns, where every instance was annotated as either a correct hypernym/hyponym pair or not. The most productive patterns (patterns 1 and 2) reached the accuracy of 56\% and 60\% respectively. Pattern 3 was much less reliable ($<$10\% accuracy) and has been allocated a lower weight. Other patterns (e.g. ``to be known/denoted as'') reached less than 5\% accuracy and have thus been excluded from the final system. The term similarity method correctly identified the correct hypernym among the top three hypernym candidates in 56\% of cases when measured on random terms from the database having at least one hypernym.

\subsection{Translation Candidates Extraction}
\label{subsec:candidates}

All entries in the TeZK terminological thesaurus are organized around the pivot language (Czech), which is also used for definitions of particular term meanings. Each term entry may contain translations to equivalent terms in five foreign languages: English, German, French, Slovak and Russian. When editing a term, the TeZK system suggests translation candidates based on the domain corpora of all these languages. Since these corpora are not parallel, i.e. they do not contain translations of documents but rather documents from the same domain, it is not possible to use standard word-alignment tools such as GIZA++\cite{Och:2003:SCV:778822.778824} to identify direct phrase translations:  the corpora are not aligned at the sentence level so standard co-occurrence statistics cannot be used.

We exploit the fact that equivalent terms share similar contexts. Consider two terms: ``European Parliament'' in English and ``Parlement europ\'{e}en'' in French. In English documents, the term co-occurs with verbs like ``revokes'', ``decided'', ``informed'', ``opposed'' and in French, ``Parlement europ\'{e}en'' co-occurs (collocates) with ``r\'{e}voque'', ``inform\'{e}'', ``notifi\'{e}'', ``oppos\'{e}''. In many cases, the collocations are translations of each other. It is possible to extract the collocates for each monolingual candidate term and measure how much these collocates overlap for all possible pairs of English and French terms using an English-French translation dictionary. It may seem paradoxical to require an existing translation dictionary to extract translation candidates but it is important to note that the dictionary does not need to contain special terminology: the collocations are usually frequent, core language items as is the case with the verbs above, thus they appear in even modest-size dictionaries. And since we limit the selection on both sides to the identified ``term candidates'', see~\ref{subsec:termextraction}, the computations are not overloaded by very frequent collocating words (e.g. frequent verbs) which have a lot of translation possibilities.

For each pair of terms, i.e. a source language (Czech) term and a candidate term in a target language, the number of collocates which are translation equivalents according to a given dictionary, is computed. These numbers are then used for ordering all pairs of terms and the top 10 candidates for each source language term are selected. The accuracy of this method is definitely not sufficient for automatic translation, thus the system uses the resulting lists as a reservoir of related terms (and thus translation candidates) in the target language. To evaluate the method on the existing data, the resulting candidate translations were compared to the existing term translations.\footnote{The current term dictionary contained from 3,070 to 4,575 translations from Czech terms to terms in the other five languages.} In 34\% of Czech terms, the existing English translation appeared in the top 20 automatically identified translation candidates. For the other languages this ratio is 40\% for German, 21\% for French, 24\% for Russian and 47\% for Slovak. These numbers do not seem adequate for helping the editor, but please note that in this evaluation possible lexical variations and synonyms were not taken into account as the annotations for them were not available. The results thus express only exact matches, but (different) possible translation candidates were often available in the first 10 items. 

Examples of correctly identified candidate translations are:
\begin{itemize}
\item \textit{katastr} (cadastre): land cover, cadastre, land information, land, land survey, land registry, land management, information system, georeference, control. 
\item \textit{m\v{e}\v{r}en\'{i}} (measurement): measurement, point positioning, dual frequency, position, fitting, distance measurement, land cover, ellipsoidal height, vertical control, vertical angle.
\item \textit{poloha} (position, location): position, location, measurement, height, ellipsoidal height, positioning, orthometric height, navigation system, ground control, accuracy.
\end{itemize}
Examples of where the method did not yield correct candidates are:
\begin{itemize}
\item \textit{katastr\'{a}ln\'{i} \'{u}\v{r}ad} (cadastral office): registration, indexing, geospatial information, geodetic datum, analysis, factor, bench mark, system, scan line, process control.
\item \textit{atlas} (atlas): input, format, control, system, conformity assessment, cadastre, raster image, quality control, monitoring, data, alteration.
\end{itemize}
For the purpose of the evaluation, which was measured on the existing translations of 2,972 to 8,439 terms (based on the respective language pair), the statistical bilingual dictionaries were built from the parallel corpus OPUS2\cite{tiedemann2009news} and the DGT-TM translation memory\cite{steinberger2013dgt}.

\section{Thesaurus Management Application}
\label{sec:application}

The main entry point of the TeZK system for the user is a web-based application accessible from all major web browsers without the need to install any new components. The application offers different modes of operation depending on the type of user. This includes 
\begin{itemize}
\item searching and browsing term information including term usage examples, term relations or the term hierarchy, 
\item term entry editing, and
\item full terminological thesaurus management with the processing of both new terms added by terminologists as well as automatically extracted new candidate terms.
\end{itemize}
The whole application is based on our general dictionary browsing and editing development platform, DEB, which is briefly presented in the following section.

\subsection{Dictionary Editor and Browser Platform}
\label{subsec:deb}

Exploiting our experience of several lexicographic projects, we have designed and implemented a universal dictionary writing system that can be exploited in lexicographic applications to build and exploit both small and large lexical databases. The system is called Dictionary Editor and Browser, or DEB.\cite{gwc2008_hales_pala_xrambous} Since 2005, DEB has been employed in more than 20 international research projects. Examples of applications based on the DEB platform include the Czech Lexical Database\cite{horak2013praled} with detailed information on more than 213,000 Czech words, or the complex lexical database, Cornetto, combining the Dutch WordNet, an ontology, and an elaborate lexicon.\cite{cicling2008} Current ongoing projects include the Pattern Dictionary of English Verbs tightly interlinked with corpus evidence,\cite{ELMAAROUF14.34} the Dictionary of Family names in Britain and Ireland\cite{fanuk} providing a detailed investigation into over 45,000 surnames to be published by Oxford University Press, and a compilation of the Dictionary of the Czech Sign Language with extensive use of multimedia recordings to present the signs visually.
The DEB platform is based on a client-server architecture, which provides a raft of benefits. All the data are stored on a server and a considerable part of its functionality is also implemented on the server, which permits the client application to be very lightweight. The server part is built from small, reusable parts, called servlets, which allow a modular composition of all services. Each servlet provides different functionalities such as database access, dictionary search, morphological analysis or connections to corpora.

The overall design of the DEB platform focuses on modularity. The data stored in a DEB server can be saved in any kind of structural database (or several different databases) and the results are combined in the answers to user queries without the need to use specific query languages for each data source. The main data storage is currently provided by the Sedna XML database,\cite{fomichev2006sedna} which is an open-source native XML database providing XPath and XQuery access to a set of document containers. 

The user interface, that forms the most important part of a client application, usually consists of a set of flexible complex forms which dynamically cooperate with the server parts. A DEB client application can be implemented in any programming language which allows interaction with the DEB server using the available server interfaces.\footnote{Client applications communicate with servlets using HTTP requests in a manner similar to a popular concept in web development called AJAX (Asynchronous JavaScript and XML) or using the W3C standard SOAP protocol. The data are transported over HTTP in a variety of formats: RDF, XML documents, JSON-encoded data, plain-text formats, or marshalled using SOAP.} Details regarding the TeZK client application are presented in Section~\ref{subsec:editing}.

The main assets of the DEB development platform are:
\begin{itemize}
\item All the data are stored on a server and a considerable part of the functionalities is also implemented on a server, allowing the client application to be very lightweight.
\item It provides very good tools for (remote) team cooperation so that data modifications are immediately seen by all users. The server also provides authentication and authorization tools.
\item A DEB server may offer different interfaces using the same data structure. These interfaces can be reused by many client applications. 
\item Homogeneity of the data structure and presentation. If an administrator commits a change in the data presentation, this change will automatically appear in every instance of the client software.
\item Easy integration with external applications via API (Application Programming Interface).
\end{itemize}

\subsection{Initial Thesaurus Data}

Although the main aim of the TeZK terminological thesaurus development lies in managing and publishing the authoritative specialized terminology and its updates both to experts in the subject field and the general public, the terminological thesaurus also contains a broad vocabulary of related terms. Users may even search for unofficial terms, and thanks to the term relations and detailed information on the source of a given term, users can easily explore all related terms and navigate to the preferred ``official'' term variant.

To build the initial TeZK terminological thesaurus data covering a broad domain vocabulary, we have combined several resources. In the first stage, the current Czech authoritative terminology dictionary\footnote{\textit{Terminologick\'{y} slovn\'{i}k zem\v{e}m\v{e}\v{r}ictv\'{i} a katastru nemovitost\'{i}} (The Dictionary of geodesy, cartography and cadastre) is published electronically at \url{http://www.vugtk.cz/slovnik} and processed by the Terminology commission of the Czech Office for Surveying, Mapping and Cadastre.} (which contained 3,937 term definitions and translations, but did not offer a taxonomy network) was combined with a hyper/hyponymic tree of 6,800 entries\footnote{Also provided by the Terminology commission of the Czech Office for Surveying, Mapping and Cadastre.} (containing hyponymic relations, but without any detailed information about terms) and by 450 candidate terms extracted from the Czech domain corpus. 

The first two resources were available in HTML with a mostly fixed entry structure, but still leaving portions of text in an unstructured format. We have thus implemented a flexible import module for the TeZK system, which is able to import both structured and unstructured data to the terminological thesaurus. The system allows the administrators to configure the formatting rules of imported data with HTML, CSV and TXT formats currently supported. After the initial parsing of the document using the formatting rules, the import module detects duplicate, misspelled or close terms, normalizes abbreviated forms, and cleans the data (e.g. correcting the punctuation). In the next step, the imported data are converted to a unified XML format for database storage. The import module was employed to combine the two resources (terminology dictionary and the hypernymic tree) resulting in combined term entries containing both detailed term information and term relations. Generally, the module enables easy expansions of the terminological thesaurus employed, e.g. in importing terms from the regularly updated Registry of Territorial Identification (RUIAN)\footnote{\url{http://www.cuzk.cz/ruian/}}.

The resulting terminological thesaurus also contains suggestions of candidate terms automatically extracted from the Czech domain corpus\footnote{As Czech plays the role of the "pivot" language in the TeZK terminological thesaurus.}. These suggestions are inserted into a separate taxonomic category, each entry including the information regarding the source and the reliability of the term. These terms are then subject to approval or disapproval by subject field specialists using the automatic candidate functionality for emplacing the term into the correct position in the taxonomy and enriching it with the foreign language terminology translations. Table~\ref{tab:tezk} shows details regarding the current size of the terminological thesaurus.

\begin{table}
\tbl{The TeZK terminological thesaurus size statistics.}
{\hspace*{2em}\begin{tabular}{lr}
\hline
 & Number of entries \\ \hline
Total entries & 8,427 \\
Hyper/hyponymic relations & 8,827 \\
Explanations provided & 4,117 \\
Entries categorized to domain & 3,905 \\
Total number of translations & 24,973 \\
English translation & 9,073 \\
German translation & 4,513 \\
Slovak translation & 3,751 \\
Russian translation & 3,068 \\
French translation & 4,568 \\ \hline
\end{tabular}\hspace*{2em}}
\label{tab:tezk}
\end{table}

\subsection{Entry Editing}
\label{subsec:editing}

The TeZK terminological thesaurus editing module is designed and implemented as a client application, with the DEB server providing the database and management backend. The editing interface is a multi-platform web application accessible in any modern browser utilizing open-source technologies\footnote{JQuery (http://jquery.com) is used for communication and SAPUI5 (\url{https://sapui5.netweaver.ondemand.com/}) libraries for the graphic interface. The client and the server communicate using a standardized interface over HTTP with the data encoded in the JSON format.}. The standardized application interface allows for an easy integration of third-party applications that can be built upon the terminological thesaurus data. The interface provides all the functions needed to work with the data (e.g. search queries, browsing the terminological thesaurus structure and detailed entry information, entry creation and updates...). Two standard remote access techniques are available supporting modern web-service standards: REST/JSON\cite{Fielding:2002:PDM:514183.514185} and WSDL\footnote{\url{http://www.w3.org/TR/wsdl/}}. One of the intended use cases is the integration into the official public Geoportal website\footnote{\url{http://geoportal.cuzk.cz/}}, where the terminology is to be used for the document metadata and categorization.

\begin{figure*}[ht]
\center
\includegraphics[width=.8\textwidth]{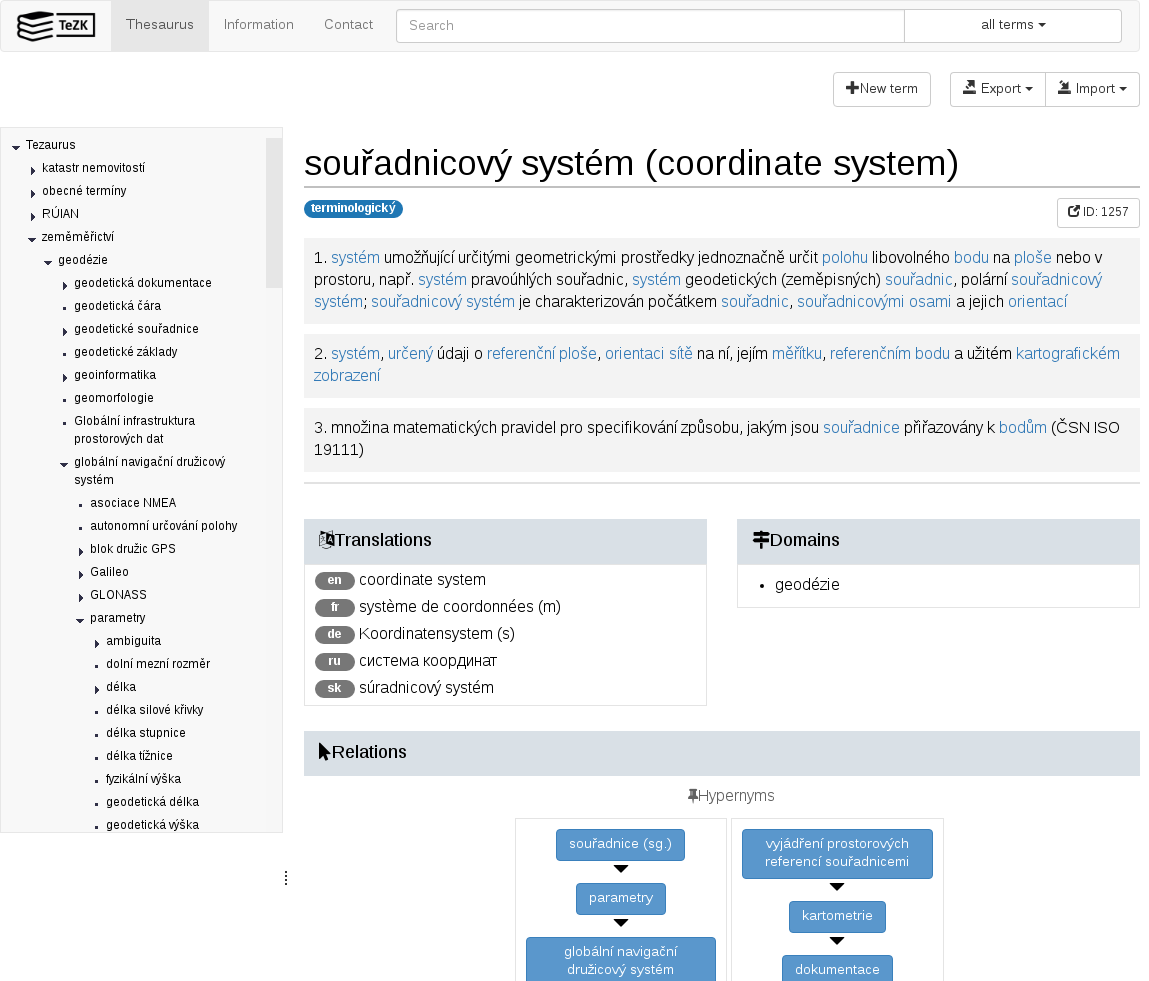}
\caption{Browsing the terminological thesaurus, with detailed information for one term.}
\label{fig:browsing}
\end{figure*}

\begin{figure*}[ht]
\center
\includegraphics[width=.8\textwidth]{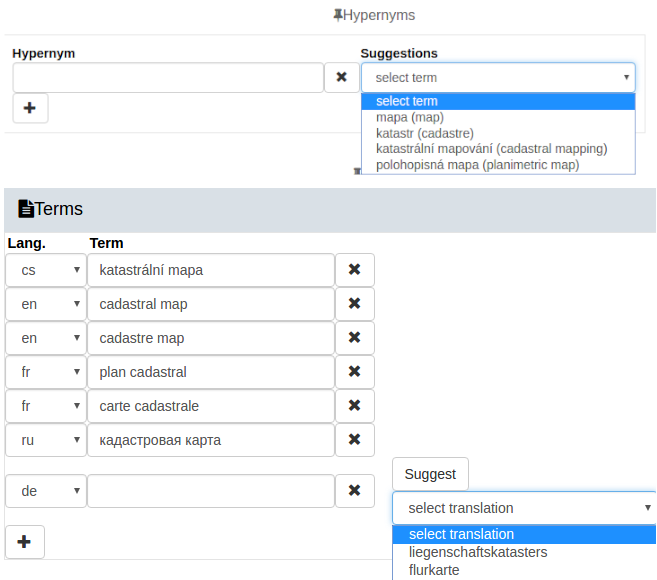}
\caption{Automatic broader candidate terms and translations offered within the editor, for the term katastr\'{a}ln\'{i} mapa (cadastre map).}
\label{fig:candidates}
\end{figure*}

\noindent
The TeZK terminological thesaurus application offers a graphical interface for browsing the hypernymic tree (see Figure~\ref{fig:browsing}). There are several possible visualization techniques for the taxonomy and the TeZK browser works with an expanding multi-level tree. Although it may not display all the relations in a proper graph form (as a term can have more than one hypernym), the expanding tree is the most intuitive representation for users. If a term has more than one hypernym, it is presented multiple times in the tree structure (with the same unique identification number). The automatic hypernym candidates, obtained via the technique described in Section~\ref{subsec:hypernym}, are offered to the editor directly within the editing form (see Figure~\ref{fig:candidates}).

Each term presentation includes all meaning explanations, translations, and accepted variants. In case of new terms, the translation candidates are selected from the TeZK domain corpora (see Section~\ref{subsec:candidates} for details) and offered as a list of proposals in the editing form (see also Figure~\ref{fig:candidates}). When more sources are incorporated into the terminological thesaurus, the reliability of each source and revision history is presented to the users. Source reliability follows the rating scale of the authoritative body---the most reliable being the terms authorized by the terminology committee, followed by terms used in scientific journals, with terms made up by the general public at the bottom of the scale. Users and third-party applications may decide which sources or terms they prefer to work with.

\begin{figure*}[t]
\center
\includegraphics[width=.8\textwidth]{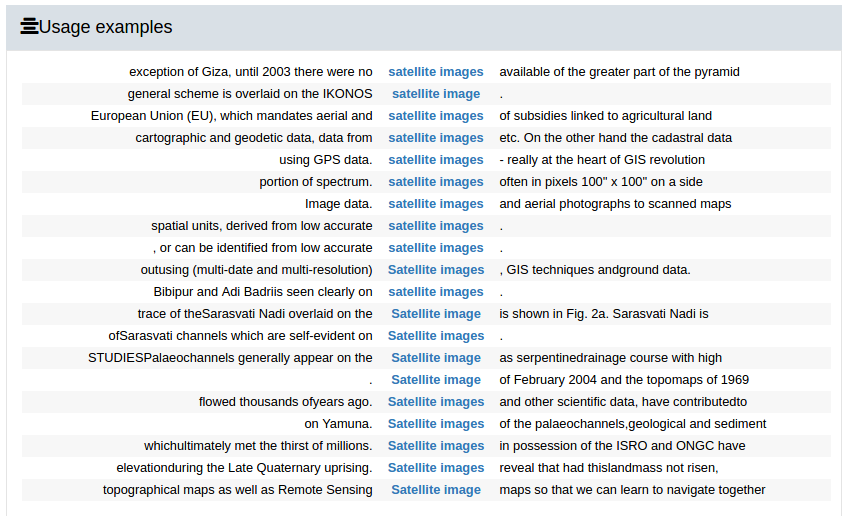}
\caption{Corpus evidence for context usage examples of the selected term \textit{dru\v{z}icov\'{y} sn\'{i}mek} (sattelite image).}
\label{fig:corpus}
\end{figure*}

\begin{figure*}[t]
\center
\includegraphics[width=.8\textwidth]{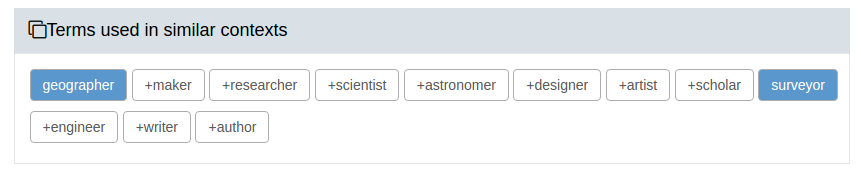}
\caption{Related words (words used in similar contexts) to the selected term---kartograf (cartographer). Words in blue boxes are already present in the TeZK terminological thesaurus.}
\label{fig:related}
\end{figure*}

Definitions created by specialists are advantageously supplemented with real-world examples of their usage in contexts based on the domain corpora (see Figure~\ref{fig:corpus}) or related terms, which are obtained by comparing the term context words and offering a list of terms used in similar contexts (see Figure~\ref{fig:related}). The ``relatedness'' of two terms is computed by comparing their collocational tables of words that appear most often in the surrounding contexts of each term.

\subsection{Linked Open Data}

The term Linked (Open) Data (LOD) refers to a methodology for publishing and interlinking remote structured data via online references. This methodology was proposed by Berners-Lee.\cite{berners2006design} The importance of Linked Open Data is acknowledged for example by the European Union, funding projects like LOD21\footnote{\url{http://lod2.eu/}} (a large integrated project to develop tools, standards and management methods for Linked Open Data) or Open Data Portal\footnote{\url{http://data.europa.eu/euodp/en/data}} (catalogue of data available for reuse). The methodology introduces five requirements for data to conform with the LOD principles, which include availability under open license, encoding in open machine-readable structured format, or interlinking with other LOD resources.

Since the Czech Cadastre Office aims to publish the data compiled in the TeZK terminological thesaurus for public use, the system needs to support the Linked Open Data methodology. The DEB platform provides the appropriate tools, but the decisions on how to release the data lie with the author. The DEB platform functionality generally enables publishing all documents as genuine Linked Open Data. All the terminological thesaurus data is accessible via standard web service protocols (WSDL/SOAP), encoded in a standardized XML structure (RDF/SKOS) interlinked to their respective sources. The only requirement outside the system is thus the decision about the (open) license for the terminological thesaurus data reuse. With data in XML format, it is possible to use various techniques for complex data querying, using for example the refinement framework implemented by Bao et al.\cite{Bao2017}

\section{Conclusions and Future Work}
\label{sec:conclusion}

We have presented the details of a new lexicographic system, which builds extensively on automatic language processing techniques to enable creating and regularly updating a domain terminology. The TeZK system offers a unique combination of a well-prepared lexicographic database system with experimental techniques exploiting information derived from large domain corpora in several languages. The system uses the corpora to provide an automated functionality for extracting terminology candidates based on the available documents as well as offering proposals to incorporate new terms into the terminological thesaurus taxonomy and supplement them with correct term translations into five foreign languages.

The terminology of the land surveying domain is maintained by the Terminology commission of the Czech Office for Surveying, Mapping and Cadastre that approves new terms and their official translations. The Commission will use the TeZK system to review automatically extracted terms and user-submitted terms. With the help of the automatic candidate selection functions, updated versions of the official terminology will be published periodically. Since each term will be rated, public users may quickly check if a term is officially recommended and find the right term for the task. The thesaurus contains knowledge-rich contextual information for each term (definitions, corpus examples, word relations), making it an inestimable resource for all kinds of works related to the specific domain. Thanks to the TeZK system support for Linked Open Data methodology and standardized public API application interface, the published resources and the terminological thesaurus functionality can be seamlessly integrated into third-party applications.

In the future work, we will further investigate the techniques for candidate translations identification. We plan to employ distributional semantics models as another measure for ordering and classifying the candidate terms in the target language.

The TeZK system will also serve as a basis for the Czech e-Government registry of terminological thesauri, currently in the early development phase. In a follow-up project, the terminological thesaurus system is being updated to support easy and user friendly deployment at any organization (both government organizations, and unofficial interest associations), with the possibility to customize work processes based on specific organization requirements. Furthermore, each instance of the terminological thesaurus system will share data with the central registry and all other terminological thesauri. During 2019, the whole system will be tested with two terminological thesauri -- a thesaurus of geospatial information terminology, and the Ministry of Interior law terminology thesaurus.

\bibliographystyle{ws-ijait}
\bibliography{thesref}

\end{document}